%% file: main.tex
\documentclass[letterpaper]{article} % DO NOT CHANGE THIS
\usepackage{cite}
\usepackage{amsmath,amssymb,amsfonts}
\usepackage{algorithmic}
\usepackage{graphicx}
\usepackage{textcomp}
\usepackage{xcolor}
\usepackage{bm}
\usepackage{multirow}
\usepackage{subfig}
\usepackage{threeparttable}
\usepackage{booktabs}
\usepackage{xspace}
\usepackage{makecell}
\usepackage{diagbox}
\usepackage{aaai24}
\usepackage{times}  % DO NOT CHANGE THIS
\usepackage{helvet}  % DO NOT CHANGE THIS
\usepackage{courier}  % DO NOT CHANGE THIS
\usepackage[hyphens]{url}  % DO NOT CHANGE THIS
\usepackage{graphicx} % DO NOT CHANGE THIS
\usepackage{natbib}  % DO NOT CHANGE THIS AND DO NOT ADD ANY OPTIONS TO IT
\usepackage{caption} % DO NOT CHANGE THIS AND DO NOT ADD ANY OPTIONS TO IT
\usepackage{algorithm}
\usepackage{algorithmic}

\usepackage{newfloat}
\usepackage{listings}
\DeclareCaptionStyle{ruled}{labelfont=normalfont,labelsep=colon,strut=off} % DO NOT CHANGE THIS
\floatstyle{ruled}
\newfloat{listing}{tb}{lst}{}
\floatname{listing}{Listing}
\newcommand\figref[1]{Figure.~\ref{#1}}

\newcommand\tabref[1]{Table~\ref{#1}}

\newcommand{\sysname}{STLinear\xspace}

\begin{document}
\title{Minimalist Traffic Prediction: Linear Layer Is All You Need}

\author{
    Wenying Duan\textsuperscript{\rm 1},
    Hong Rao\textsuperscript{\rm 1},
    Wei Huang\textsuperscript{\rm 1},
    Xiaoxi He\textsuperscript{\rm2}\thanks{corresponding author}
}
\affiliations{
    \textsuperscript{\rm 1}Nanchang University, Nanchang, China\\
    \textsuperscript{\rm 2}University of Macau, Macau\\
    \{wenyingduan, raohong, huangwei\}@ncu.edu.cn, hexiaoxi@um.edu.mo
}

\maketitle

\begin{abstract}
Traffic prediction is essential for the progression of Intelligent Transportation Systems (ITS) and the vision of smart cities. While Spatial-Temporal Graph Neural Networks (STGNNs) have shown promise in this domain by leveraging Graph Neural Networks (GNNs) integrated with either RNNs or Transformers, they present challenges such as computational complexity, gradient issues, and resource-intensiveness. This paper addresses these challenges, advocating for three main solutions: a node-embedding approach, time series decomposition, and periodicity learning. We introduce \sysname, a minimalist model architecture designed for optimized efficiency and performance. Unlike traditional STGNNs, \sysname operates fully locally, avoiding inter-node data exchanges, and relies exclusively on linear layers, drastically cutting computational demands. Our empirical studies on real-world datasets confirm \sysname's prowess, matching or exceeding the accuracy of leading STGNNs, but with significantly reduced complexity and computation overhead (more than 95\% reduction in MACs per epoch compared to state-of-the-art STGNN baseline published in 2023). In summary, \sysname emerges as a potent, efficient alternative to conventional STGNNs, with profound implications for the future of ITS and smart city initiatives.
\end{abstract}

\input{body/introduction}
\input{body/related}
\input{body/method}

\input{body/experiments}

\input{body/conclusion}

\clearpage
\bibliography{cites}
\end{document}

%% file: body/introduction.tex
\section{Introduction}

Traffic prediction is fundamental to Intelligent Transportation Systems (ITS), which aim to bolster the efficiency and safety of urban traffic management. By leveraging historical data, traffic flow prediction strives to forecast future traffic conditions. Precise predictions can drive the growth of smart cities by curbing traffic congestion, cutting fuel consumption, reducing greenhouse gas emissions, and minimizing road accidents. A key characteristic of traffic data is the strong correlation among time series due to the interconnectedness of the road network. For precise traffic prediction, numerous state-of-the-art approaches \cite{DBLP:conf/ijcai/YuYZ18,bib:IJCAI19:Wu,bib:IJCAI20:Huang,DBLP:conf/aaai/ZhengFW020,DBLP:conf/aaai/LiZ21,bib:AAAI22:Choi,DBLP:conf/aaai/JiangHZW23} employ Spatial-Temporal Graph Neural Networks (STGNNs).

Spatial-Temporal Graph Neural Networks (STGNNs) are tailored for processing spatial-temporal data, merging two essential components: a spatial module and a temporal module. The spatial module harnesses the power of Graph Neural Networks (GNNs) to model spatial information, capturing intricate relationships between entities. Conversely, for temporal information, both Recurrent Neural Networks (RNNs) \cite{connor1992recurrent,DBLP:conf/iclr/LiYS018,DBLP:conf/nips/0001YL0020} and Transformers \cite{vaswani2017attention, DBLP:conf/aaai/JiangHZW23,DBLP:conf/icde/0001LGWZSHW23} have been utilized in STGNN research. This union of either RNNs or Transformers with GNNs supercharges STGNNs for tasks like traffic prediction. Nonetheless, STGNNs present their own challenges:

\begin{itemize}
    \item \textbf{Complexity of GNNs:} While GNNs are undoubtedly effective, they aren't optimized in terms of resource utilization. Their integration into the STGNN framework amplifies both the computational and structural complexities. This can lead to increased resource demands and extended training periods. As an illustrative example, consider the adaptive graph convolution network (AGCN), a key advancement within STGNNs. The inference time complexity of a single AGCN layer is denoted by $\mathcal{O}(N^{2})$, where $N$ represents the number of nodes \cite{10.1145/3580305.3599418}. 
    \item \textbf{RNNs and Gradient Issues:} While RNNs excel at modeling short-term dependencies, they are susceptible to gradient explosion and vanishing, especially with extended data sequences. This not only affects prediction accuracy but also complicates training, often requiring specialized techniques or interventions.
    \item \textbf{Resource Intensive Transformers:} Although Transformers excel at handling long-term dependencies due to their multi-head self-attention mechanism, they can be resource-intensive, especially on large datasets, making them less suitable for certain real-time or resource-restricted applications.
\end{itemize}

In summary, despite the successes of STGNN-based techniques, they grapple with challenges like growing complexity and marginal improvements, especially for long-term traffic prediction across many nodes. This raises the question: \textit{Is it possible to develop a model as proficient as STGNNs but more efficient in both training and inference?}
Given recent advancements in spatial-temporal data processing research, we have pinpointed potential solutions for STGNNs' efficiency constraints:
\begin{enumerate}
    \item Recent findings \cite{10.1145/3580305.3599418} suggest that temporal dependency is paramount for inference in many spatial-temporal tasks. It suggests that current STGNN structures might overemphasize spatial dependency modeling, leading to increased overhead and reduced scalability. Nonetheless, the inherent spatial attributes of individual nodes are vital \cite{DBLP:conf/nips/0001YL0020}, particularly during cooperative training processes. We advocate for a node-embedding approach, analogous to the soft weight sharing in multi-task learning, enabling parameter extraction from a unified weight pool.
    \item The often low information density and abundant noise in traffic data highlight the need to recognize time-dependencies across varied scales. Recent studies \cite{DBLP:conf/aaai/ZengCZ023,DBLP:journals/corr/abs-2302-04501, 10.1145/3580305.3599533} propose multi-scale information abstraction as the chief method for spatial-temporal inference. They indicate that basic linear models combined with decomposition might rival or even surpass Transformers. In response, we endorse time series decomposition.
    \item Traffic systems usually reflect the periodicity of human society. 
    Many works have shown that considering periodicity is beneficial for traffic prediction \cite{DBLP:conf/aaai/ZhouZPZLXZ21, DBLP:conf/aaai/JiangHZW23}.
    Recognizing the crucial role of periodic information, we present periodicity learning, an innovative mechanism to encode and smoothly integrate periodicity data into our processes.
\end{enumerate}

Culminating our efforts, we propose \textbf{\sysname}, a minimalist design of model architecture providing state-of-the-art performance and efficiency. \sysname is fully localized, meaning it requires no inter-node data exchange during training and inference. This not only cuts down the model's computational overhead but also facilitates its deployment on systems with communication constraints. Additionally, \sysname operates on pure linear layers, drastically reducing computational resource needs and simplifying both training and inference. The architecture is depicted in \figref{fig:architecture}, with comprehensive explanations in section Method.

Our empirical assessments, carried out on four publicly available real-world datasets, demonstrate that \sysname consistently matches or even outperforms leading STGNNs in traffic prediction tasks in inference accuracy, all with significantly reduced computing overhead. For instance, when compared against SSTBAN \cite{DBLP:conf/icde/0001LGWZSHW23}, a benchmark state-of-the-art method, \sysname records a remarkable reduction in MACs per epoch during inference, ranging from 95.50\% to 99.81\%. During training, this efficiency becomes even more evident: \sysname exhibits a reduction in MACs per epoch between 99.10\% and 99.96\% and a decrease in memory usage that varies between 15.99\% and 95.20\%.

%% file: body/related.tex
\section{Related Work}

Our work majorly relates to two research fields: \textit{(i)} traffic prediction using STGNNs, and \textit{(ii)} long time series forecasting (LTSF) with linear models.

\subsection{STGNNs for Traffic Prediction}
Traffic forecasting is crucial within ITS and has been a subject of study for many years. Historically, techniques like ARIMA \cite{makridakis1997arma} and VAR \cite{zivot2006vector} centered on temporal dependencies for individual nodes or a small group of nodes. However, these approaches struggled with real-world traffic data that exhibited both temporal and spatial dynamics. 
The advent of deep learning brought to the fore the STGNNs, which have become indispensable in this domain. 
Their strength lies in unveiling hidden patterns of spatially irregular signals over time. 
Typically, these models combine graph convolutional networks (GCNs) with sequential models (e.g., RNNs, TCN, Transformer). Below, we detail the design components of prevalent STGNN-based traffic prediction methods.

\subsubsection{Spatial Dependency Modeling}
A central challenge for STGNN techniques is accurately modeling the intricate spatial dependencies between nodes, represented by the adjacency matrix. While GCNs \cite{DBLP:conf/ijcai/YuYZ18, DBLP:conf/aaai/GuoLFSW19,DBLP:conf/kdd/PanLW00Z19} are a popular choice for spatial dependency modeling, they often rely on a predetermined graph structure which might not fully encapsulate spatial dependencies \cite{DBLP:conf/nips/0001YL0020}. Graph WaveNet \cite{bib:IJCAI19:Wu}  proposed an adaptive graph convolutional (AGCN) layer that learns an adaptive adjacency matrix, bypassing the need for a preset graph. AGCRN \cite{DBLP:conf/nips/0001YL0020} went a step further, positing that adaptive graphs alone might not be enough. This method enriches AGCN with a node adaptive parameter learning (NAPL) module, which discerns node-specific patterns using node embedding. Inspired by NAPL, we endorse a node-embedding method, akin to the soft weight sharing in multi-task learning. This allows for parameter extraction from a singular weight repository. It's worth noting that AGCRN and its derivatives \cite{bib:AAAI22:Choi,bib:ICLR22:Chen} might overemphasize spatial dependency modeling, leading to elevated overhead and diminished scalability. 
Distinctly, \sysname emphasizes spatial characteristics rather than sheer spatial dependencies.

\subsubsection{Temporal Dependency Modeling}
Being a unique kind of time series, STGNNs must also grapple with capturing temporal dependencies. Based on the methodologies employed to tackle time dependencies, STGNNs can be categorized into RNN-based, CNN-based, and Transformer-based models \cite{DBLP:conf/iclr/LiYS018, DBLP:conf/nips/0001YL0020, bib:ICML21:Chen, bib:ICLR22:Chen}. Graph Wavenet, STSGCN, and ASTGCN stand out as CNN-based methods, whereas GMAN \cite{DBLP:conf/aaai/ZhengFW020}, PDFormer \cite{DBLP:conf/aaai/JiangHZW23}, and SSTBAN \cite{DBLP:conf/icde/0001LGWZSHW23} are prominent Transformer-based techniques. Recent literature underscores the benefits of incorporating historical periodicity information for temporal dependency modeling. The straightforward approach involves injecting learnable daily and weekly periodicity embeddings into the input sequence. Contrary to this approach, \sysname incorporates the periodicity embeddings of only the initial and terminal data points in the historical dataset, optimizing for efficiency.

\subsection{LTSF with linear models}
Linear models have witnessed a resurgence in the LTSF domain \cite{DBLP:conf/cikm/ShaoZ00X22,DBLP:conf/aaai/ZengCZ023}. For instance, LightTS leverages linear models alongside two nuanced downsampling strategies to enhance forecasting \cite{zhang2022less}. After deconstructing the time series into trend and remainder components, DLinear applies a simple linear model, even outshining Transformer-based state-of-the-art solutions \cite{DBLP:conf/aaai/ZengCZ023}. TSMixer, designed around multi-layer perceptron (MLP) modules, targets multivariate long time series forecasting, presenting an efficient alternative to Transformers \cite{10.1145/3580305.3599533}. 
Yet, these models falter with traffic data's unique characteristics. In contrast, \sysname adeptly incorporates periodicity and spatial attributes.

%% file: body/method.tex
\section{Efficient Traffic Prediction with STLinear}

\begin{figure}[t]
    \centering
    \includegraphics[width=1\linewidth]{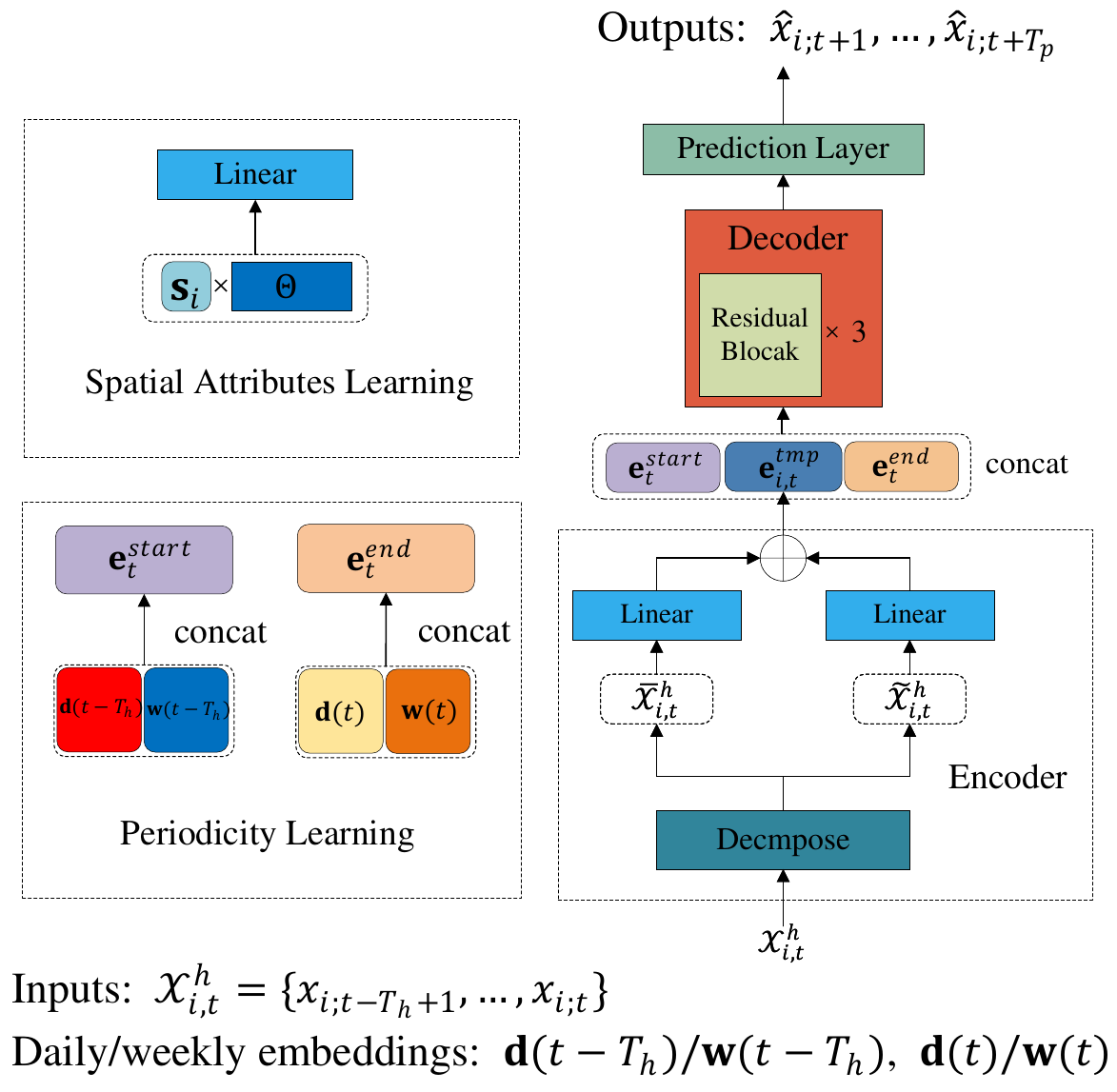}
    \caption{Overview of STLinear's architecture}
    \label{fig:architecture}
\end{figure}

This section introduces \sysname, a novel method for traffic prediction that emphasizes efficiency in both training and inference stages. Refer to Figure \ref{fig:architecture} for an overview of our architecture.

\subsection{Notation and Problem Definition}

Adhering the conventions in STGNNs-enabled traffic forecasting \cite{bib:IJCAI19:Wu,DBLP:conf/nips/0001YL0020, bib:ICLR22:Chen}, traffic data is denoted as  $\mathcal{X} \in\mathbb{R}^{N \times T\times C}$, where $N$ stands for the number of nodes, $T$ the number of time-steps, and $C$ the number of features.
A significant portion of traffic data, especially those explored in this paper, are univariate, meaning $C = 1$. Thus, for simplicity, we employ $\mathcal{X} \in\mathbb{R}^{N \times T}$ without any loss of generality.

$\mathcal{X}$ can be depicted as $\{\mathbf{x}_{:,1}, \mathbf{x}_{:,2}, \ldots, \mathbf{x}_{:,T}\}$ with $\mathbf{x}_{:,t} \in \mathbb{R}^{N}$ being a “snap shot" of all nodes' feature at time $t$, or as $\{\mathbf{x}_{1,:}, \mathbf{x}_{2,:}, \ldots, \mathbf{x}_{N,:}\}$ with $\mathbf{x}_{i,:} \in \mathbb{R}^{T}$ being the time series data of the $i$-th node across all time-steps.
In this paper, we make use of both forms of representations. 

% Following the conventions in traffic forecasting using STGNNs researches \cite{bib:IJCAI19:Wu,DBLP:conf/nips/0001YL0020, bib:ICLR22:Chen}, we consider the traffic data that contains $N$ univariate time series represented as $\mathcal{X}=\{{X}_{:,1}, {X}_{:,2}, \ldots, {X}_{:,T}\}$, $\mathcal{X} \in\mathbb{R}^{N \times T\times C}$, $N$ is also known as the number of nodes in traffic data,
% is the time series for the $i$-th node.
% In practice, $C=1$, so we have $\mathcal{X}_i\in \mathbb{R}^{T}$.

In traffic prediction tasks, we aim to forecast traffic for multiple steps ahead.
At time $t$, given $T_h$ historical observations symbolized as $\mathcal{X}^h_t = \{\mathbf{x}_{:,t-T_h+1} \ldots, \mathbf{x}_{:,t}\} \in\mathbb{R}^{N \times T_h}$, the task is to learn a function $\mathcal{F}(\cdot)$ mapping the historical observations into the future observations in the upcoming $T_p$ time-steps:
\begin{equation}
\left\{\mathbf{x}_{:,t+1}, \ldots, \mathbf{x}_{:,t+T_p}\right\} = \mathcal{F}(\mathcal{X}^h_t)
\end{equation}

\subsection{Linear Encoder}
\subsubsection{Time Series Decomposition.} 
Recent studies have demonstrated that simple linear models combined with decomposition techniques can achieve competitive or even superior performance against transformers on LSTF tasks \cite{DBLP:conf/aaai/ZengCZ023,DBLP:journals/corr/abs-2302-04501, 10.1145/3580305.3599533}. 
Inspired by these findings, we incorporate a time series decomposition method to enhance our temporal dependency modeling.

For the $i$-th node, the $T_h$ historical observations at time-step $t$ are denoted as $\mathcal{X}^h_{i,t} = \{{x}_{i,t-T_h+1} \ldots, {x}_{i,t}\} \in \mathbb{R}^{T_h}$, which is further decomposed into a trend component $\mathcal{\overline{X}}^h_{i,t}\in \mathbb{R}^{T_h}$ and a remainder component $\mathcal{\tilde{X}}^h_{i,t}\in \mathbb{R}^{T_h}$ using a moving average kernel. 
These components are subsequently processed using two linear layers and a summation operation to compute the temporal embedding:
\begin{align}
% \begin{gathered}
\overline{\mathbf{e}}_{i,t} &= \mathbf{W}_{tr} \cdot \mathcal{ \overline{X}}^h_{i,t} +\mathbf{b}_{{tr}} \label{eq:decompose1}\\
\tilde{\mathbf{e}}_{i,t} &= \mathbf{W}_{re} \cdot \mathcal{\tilde{X}}^h_{i,t} +\mathbf{b}_{{re}} \label{eq:decompose2}\\
\mathbf{e}^{tmp}_{i,t} &= \overline{\mathbf{e}}_{i,t} + \tilde{\mathbf{e}}_{i,t} \label{eq:temp_emb}
% \end{gathered}
\end{align}
In these equations, $\overline{\mathbf{e}}_{i,t}$, $\tilde{\mathbf{e}}_{i,t}$, and $\mathbf{e}^{tmp}_{i,t}$ all possess a dimension of $\mathbb{R}^d$.
Here, $d$ is a modifiable hyperparameter dictating the volume of encoded information.
$\mathbf{W}_{tr}, \mathbf{W}_{re}\in \mathbb{R}^{d \times T_h}$ and $\mathbf{b}_{tr}, \mathbf{b}_{re}\in \mathbb{R}^{d}$ are the learnable model weights and biases of the linear layers.
 
\subsubsection{Spatial Attributes Learning.} In Equations \eqref{eq:decompose1} and \eqref{eq:decompose2}, the learnable parameters $\mathbf{W}_{tr}$,  $\mathbf{W}_{re}$, $\mathbf{b}_{tr}$, and $\mathbf{b}_{re}$ are indeed $i$-invariant, i.e., shared between all nodes. However, traffic data from various nodes often manifest diverse patterns. Solely sharing parameters among nodes does not address the unique characteristics of each node based on their spatial locations, which can result in suboptimal predictions.

Exclusively assigning distinct sets of parameters to every node, however, might not be the optimal solution either.
Research has shown that knowledge sharing across nodes can enhance both inference accuracy and efficiency \cite{DBLP:conf/nips/0001YL0020}. Given the limited data available for each node, sharing knowledge among nodes—akin to multitask learning paradigms—can elevate the overall inference accuracy.

To strike a balance between these two aspects, we adjust Equations \eqref{eq:decompose1} and \eqref{eq:decompose2} by introducing a spatial embedding vector $\mathbf{s}_{i} \in \mathbb{R}^{e}$ for each node:
\begin{align}
\overline{\mathbf{e}}_{i,t} &= (\Theta_{tr} \mathbf{s}_{i}) \mathcal{ \overline{X}}^h_{i,t} +\mathbf{\beta}_{tr}  \mathbf{s}_{i} \\
\tilde{\mathbf{e}}_{i,t} &= (\Theta_{re} \mathbf{s}_{i})  \mathcal{ \overline{X}}^h_{i,t} +\mathbf{\beta}_{re}  \mathbf{s}_{i}
\end{align}
Here, $\Theta_{{tr}}, \Theta_{{re}}\in \mathbb{R}^{d \times T_h \times e}$ and $\beta_{tr}$, $\beta_{re}\in \mathbb{R}^{d \times e}$ serve as learnable parameter pools.
The hyperparameter $e$ decides the total number of parameters in these pools.
Effective model parameters for each node are derived by multiplying the corresponding embedding $\mathbf{s}_{i}$ with the parameter pools, extracting the node-specific weights $\Theta_{tr} \mathbf{s}_{i}, \Theta_{re} \mathbf{s}_{i} \in \mathbb{R}^{d \times T_h}$ and biases $\mathbf{\beta}_{tr}  \mathbf{s}_{i}, \mathbf{\beta}_{re}  \mathbf{s}_{i} \in \mathbb{R}^d$.
Importantly, this computation is only performed post-training, ensuring no additional computational load during inference. 
Moreover, these node-specific spatial embeddings encapsulate the \textit{spatial attributes} yet bypassing the need to model inter-node \textit{spatial dependencies}. 
This frees the model completely from the necessity of inter-node data-exchange during inference.

\subsubsection{Periodicity Learning.} Urban activities heavily influence traffic data, resulting in discernible periodicity patterns such as morning and evening rushes. To capture time-of-the-day and day-of-the-week patterns, we introduce two sets of learnable vectors: $\mathbf{d}_1, \dots, \mathbf{d}_{N_d} \in \mathbb{R}^{c}$ (where $N_{d}$ represents the number of time-steps per day) and $\mathbf{w}_1, \dots, \mathbf{w}_7 \in \mathbb{R}^{c}$. The hyperparameter $c$ is used for controlling the precision of the encoding. For a time-step $t$, the corresponding vector is denoted as $\mathbf{d}(t)$ and $\mathbf{w}(t)$,  respectively.

% $\mathbf{X}_{d(t)} $ and $\mathbf{X}_{w(t)}$ represent: the time-of-the-day and the day-of-the-week as features that can help the model learn: the cyclic patterns in the traffic data.

Leveraging the two sets of vector which encode time-of-the-day and day-of-the-week information, we passes the initial and the terminal time of a given input $\mathcal{X}^h_{i,t}$ to the model using the initial time vector and the terminal time vector
\begin{align}
    \mathbf{e}^{start}_{t} &=  [\mathbf{d}(t-T_h);\mathbf{w}(t-T_h)] \label{eq:start_emb}\\
    \mathbf{e}^{end}_{t} &= [\mathbf{d}(t);\mathbf{w}(t)]  \label{eq:end_emb}
\end{align} 
Here, $[\cdot;\cdot]$ denotes vector concatenation.

% where  $\mathbf{X}_{d,(t-\mathcal{T})}$ and $\mathbf{X}_{w,(t-\mathcal{T})}$ are the daily embedding and weekly embedding of the initial data point $\boldsymbol{X}_{i,t-\mathcal{T}}$ in $\mathcal{X}^{\mathcal{T}}$, respectively. 
% $\mathbf{X}_{d,(t-1)}$ and  $\mathbf{X}_{w,(t-1)}$ are the daily embedding and weekly embedding of the terminal data point $\boldsymbol{X}_{i,t-1}$ in $\mathcal{X}_{i}^{\mathcal{T}}$, respectively.

\subsubsection{Combining Embeddings.}
Finally, integrating all the aforementioned embedding vectors \eqref{eq:temp_emb}, \eqref{eq:start_emb}, and \eqref{eq:end_emb} which encode temporal information and spatial attributes, we derive the final embedding vector:
\begin{equation}
\label{eq:data_embedding}
    \mathbf{e}_{i,t} = [\mathbf{e}^{start}_{t}; \mathbf{e}^{tem}_{i,t}; \mathbf{e}^{end}_{t}] \in \mathbb{R}^{{(d+4c)}}
\end{equation}
% Key features of our embedding layer include:
% \begin{enumerate}
%     \item 
%     \item We apply the temporal embeddings only to the initial and terminal data-points in historical data, rather than all data-points, which decreases the computational cost.
%     \item Due to the time span of the task being less than one month, we do not introduce the monthly embedding to periodicity learning.
% \end{enumerate}

\subsection{Linear Decoder}
Our decoder employs a multi-layer structure featuring residual blocks. For the $l$-th layer:
% $\mathbf{X}^{l}_{i}=\mathrm{Res}(\mathbf{X})^{l-1}_{i}$, where $\mathbf{X}^{l-1}_{i} = \mathbf{X}_{i}$.
% The $\mathrm{ResidualBlock}(\cdot)$ is formalized as:
\begin{align}
    \mathrm{Fc}(\mathbf{y}^{l-1}_{i,t}) &= \mathbf{W}_B^l(\sigma(\mathbf{W}_A^{l}\mathbf{y}^{l-1}_{i,t}+\mathbf{b}_A^{l}))+\mathbf{b}^l_B\\
    \mathrm{Res}( \mathbf{y}^l_{i,t}) &=   \mathrm{Fc}(\mathbf{y}^{l-1}_{i,t})+\mathbf{y}^{l-1}_{i,t}
\end{align}
In the above, $\mathbf{y}^{l}{i,t}$ represents the layer output of the $l$-th residual block of node $i$ at time-step $t$.
The output of a decoder with $L$ residual blocks is denoted therefore as $\mathbf{y}_{i,t}^L$.
We also define $\mathbf{y}^{0}_{i,t} = \mathbf{e}_{i,t}$, which is the encoder output in \eqref{eq:data_embedding}. 
$\mathbf{W}_A^l$, $ \mathbf{W}_B^l$, $\mathbf{b}^l_A$, and $\mathbf{b}^l_B$ are learnable parameters for fully connected layers and are shared among nodes. $\sigma(\cdot)$ is the activation function, for which we primarily use Gaussian Error Linear Unit (GELU) in this paper.

% The encoder takes the input $\mathbf{X}_{i}$ and extracts features for prediction. The output of the encoder is denoted as: $\mathbf{H}_{i}$.
To make $T_p$-step predictions, we incorporate one extra linear layer as the output layer:
\begin{equation}
\left\{{\hat{x}}_{i,t+1}, \ldots, {\hat{x}}_{i,t+T_p}\right\} =\mathbf{W}_p \mathbf{y}_{i,t}^L +\mathbf{b}_p
\end{equation}
The final output $\left\{{\hat{x}}_{i,t+1}, \ldots, {\hat{x}}_{i,t+T_p}\right\} \in \mathbb{R}^{T_p}$ serves as the prediction of $\left\{{x}_{i,t+1}, \ldots, {x}_{i,t+T_p}\right\}$. 

\begin{table}[htb]
\centering
\small
\caption{Summary of datasets used in experiments.}
\label{tab:datasets}
\begin{tabular}{lcc}
    \toprule
     Datasets&\#Nodes& Range\\
    \midrule
     PEMS03&358&09/01/2018 - 30/11/2018\\ 
     PEMS04&307&01/01/2018 - 28/02/2018\\
     PEMS07&883&01/07/2017 - 31/08/2017\\
     PEMS08&170&01/07/2016 - 31/08/2016\\
     \bottomrule
\end{tabular}
\end{table}

%% file: body/experiments.tex
\section{Evaluations}
\label{sec:exp}
\subsection{Datasets and metrics}
\begin{table*}[t]
\centering
\caption{Overall prediction performance of different methods on the PEMS03, PEMS04, PEMS07 and PEMS08 dataset, the best results are highlighted in bold. (smaller value means better performance)}
\label{tab:results}
\tiny
\begin{tabular}{l|l|ccc|ccc|ccc|ccc}
\Xhline{1pt}
\multirow{2}{*}{Dataset} & Horizon  & \multicolumn{3}{c|}{12}   & \multicolumn{3}{c|}{48} & \multicolumn{3}{c|}{192} & \multicolumn{3}{c}{288}    \\ \cline{2-14}   
                         & Method   & MAE&RMSE&MAPE  &MAE&RMSE&MAPE       &MAE&RMSE&MAPE         &MAE&RMSE&MAPE             \\
                         \hline 
\multirow{5}{*}{PEMS03}  &DLinear  &21.18&37.67&20.22\% &27.16&60.71&28.13\% &39.20&59.98&28.22\% &38.64&58.72&28.13\% \\
                         &GraphWavenet &19.85&32.94&19.31\% &23.14&37.40&21.64\% &25.00&42.71&23.89\% &24.89&42.74&23.76\% \\
                         & DCRNN    & 17.99&30.31&18.34\% &28.07&45.00&27.93\%     
                         &28.36&45.76&26.56\% &29.69&48.03&26.79\%\\
                         & AGCRN & 16.03&28.52&14.65\% &19.68&37.29&22.38\%  
                         &22.27&41.62&25.03\% &23.49&43.15&25.29\%\\
                         
                         &STGNCDE &15.47&28.09& 15.76\% &22.36&39.92&21.66\% &23.56&41.73&22.87\% &23.93&44.93&22.91\%        \\

                        %\cline{2-14} 
                         &GMAN   & 15.11& 28.63&15.72\%   &20.45& 33.85& 20.98\% 
                         &23.98&41.35&22.81\%&N/A&N/A&N/A \\
                         &SSTBAN   &N/A&N/A&N/A  &N/A&N/A&N/A   &N/A&N/A&N/A  &N/A&N/A&N/A        \\
                         &PDFormer &14.79&25.40&15.34\% &19.89&32.75&19.59&N/A&N/A&N/A &N/A&N/A&N/A\\
                         &STID &15.20&\textbf{25.20}&16.09\%&19.82&34.80&21.06\% &22.22&40.96&23.50\% &23.01&{42.34}&25.33\%\\
                        \cline{2-14} 
                         & STLinear& \textbf{14.73}&{25.33}&\textbf{14.62\%}&\textbf{19.22}&\textbf{32.03}& \textbf{20.00\%}&\textbf{21.81}&\textbf{40.51}&\textbf{22.33\%} &\textbf{22.49}&\textbf{41.73}&\textbf{22.53\%}       \\
                         \hline
\multirow{5}{*}{PEMS04}  &DLinear  &28.95&47.31&21.22\% &39.29&60.71&28.13\% &39.20&59.98&28.22\% &38.64&58.72&28.13\% \\
                         &GraphWavenet &22.24&30.59 &16.51\% &26.40&40.60 &18.99\% &28.63&44.36&19.98\% &29.17&44.27&20.01\% \\
                         &DCRNN    & 19.71& 31.43& 13.54\%     &23.70&36.42&15.23\%   &26.78&41.40&16.10\%   &27.24&40.85&16.21\%          \\
                         &AGCRN    & 19.83& 32.30& 12.97\%    &24.18&38.26&16.31\%   &26.65&41.89&16.33\%   &26.73&42.26&16.14\%       \\
                        
                        &STGNCDE &19.21&31.09& 12.76\%   &23.79&37.62&16.01\%        &27.37&41.88&17.81\%     &29.19&42.62&18.01\%          \\

                        %\cline{2-14} 
                        
                         &GMAN   & 19.14& 31.60&13.20\%   & 23.35& 47.85& 17.98\%    &25.02& 50.68 & 18.93\%   & 25.35& 50.85& 17.74\%   \\
                         &SSTBAN   &22.63&38.24&14.69\%  &21.66& 35.51& 15.90\%    &N/A&N/A&N/A
                         &N/A&N/A&N/A          \\
                         &PDFormer &18.32&29.97&11.87\%    &21.19&34.67&14.46\%      &N/A&N/A&N/A   &N/A&N/A&N/A\\
                          &STID  &18.29&\textbf{29.82}&12.49\% &21.04&34.04&14.51\% &23.20&38.51&16.44\% &23.78&39.99&16.79\%\\
                        \cline{2-14} 
                         &STLinear &\textbf{18.21}&{29.89}&\textbf{11.87\%} &\textbf{20.54}&\textbf{33.97}& \textbf{13.66\%}
                         &\textbf{22.32}&\textbf{38.13}& \textbf{15.17\%}&\textbf{23.24}&\textbf{39.87}& \textbf{15.66\%}\\
                         \hline 
\multirow{6}{*}{PEMS07}  &DLinear  &30.29&49.63&14.36\% &42.17&63.45&22.51\% &42.09&63.14&22.43 &41.05&63.28&22.07\% \\
                         &GraphWavenet &20.65&33.49&8.91\% &24.16&40.19&10.34\% &26.82&45.73&11.58\% &26.91&46.69& 11.44\%\\
                         &DCRNN    &25.22&38.61&11.82\%  &27.48&41.31&12.53\%  &30.23&45.86&13.73\%       &34.26&53.20&15.38\%          \\
                         &AGCRN    &22.37&36.55&9.12\%   &26.10&42.78&12.03\%      &30.28&49.53&14.20\%   &31.10&49.78&14.36\%          \\
                         
                          &STGNCDE &20.53&33.84& 8.80\%  &30.34&48.75&13.33\%       &35.83&55.79&14.97\% &36.64&56.75&15.13\%          \\
                       
                        % \cline{2-14} 
                         
                         &GMAN  &20.43&33.30&8.69\% &23.78&40.15&10.25\%& 25.87&46.45&11.78\%&N/A&N/A&N/A          \\
                         &SSTBAN  &N/A&N/A&N/A  &N/A&N/A&N/A    &N/A&N/A&N/A  &N/A&N/A&N/A         \\
                         &PDFormer &19.83 &32.87&8.40\%   &23.25&41.14& 10.08\%   &N/A&N/A&N/A    & N/A&N/A&N/A     \\
                          &STID  &19.54&32.85&8.25\% &23.49&40.05&10.52\% &26.14&48.87&11.92\% &26.50&48.10&11.97\%\\
                         \cline{2-14} 
                         & STLinear&\textbf{19.19}&\textbf{32.47}&\textbf{8.07\%}  &\textbf{22.63}&\textbf{39.05}&\textbf{9.73\%}
                         &\textbf{25.18}&\textbf{44.56}&\textbf{11.57\%}&\textbf{25.57}&\textbf{45.89}&\textbf{11.41\%}\\
\hline
\multirow{6}{*}{PEMS08}  &DLinear  &21.41&30.34&15.95\% &31.99&46.17&22.55\% &32.38&47.28&22.69\% &32.99&47.39&22.63\% \\
                         &GraphWavenet &16.63&26.03&10.42\% &18.59&30.27&11.63\% &18.73&31.24&13.95\% &19.04&31.38&13.97\% \\
                         &DCRNN    &18.19&28.18&11.24\%  &18.60&30.04&12.66\%  &22.82&35.67&14.73\%       &23.27&36.94&15.10\%          \\
                         &AGCRN    &15.95&25.22& 10.09\%   &17.45&28.05&11.25\%      &21.17& 32.86&14.83\%   &21.96&33.41&12.98\%          \\
                        
                        &STGNCDE &15.45&24.81& 10.18\% &21.18&33.46&13.69\%  &23.28&37.13&14.91\%     &24.57&38.56&15.60\%          \\
                       
                         %\cline{2-14} 
                         
                         &GMAN   &15.31&24.92&10.13\% &18.70& 35.89&16.81\%&20.21&35.89&18.66\%& 20.86&40.84&18.82\%          \\
                         &SSTBAN  &18.37&24.87&10.59\%  &16.94&28.82& 12.47\% &20.07&32.32&14.29\%   &20.34&33.40&14.20\%           \\
                         &PDFormer & \textbf{13.58} & 23.51& \textbf{9.05}\%&16.91&28.28&11.74\% &18.61&31.17&13.72\% &19.63&32.57&14.19\%    \\
                         &STID  &14.20&23.49&9.28\%  &16.89&27.02&11.67\%  &18.86&31.24&13.30\% &19.14&31.94&13.81\% \\
                         \cline{2-14} 
                         & STLinear&13.95&\textbf{23.39}&9.21\%&\textbf{16.52}&\textbf{27.63}&\textbf{10.87\%}&\textbf{17.97}&\textbf{30.37}&\textbf{12.47\%} &\textbf{18.29}&\textbf{31.26}&\textbf{12.87\%}
\\
\Xhline{1pt}
\end{tabular}
\end{table*}

To assess the performance of the proposed \sysname, we carried out comprehensive experiments using four widely recognized real-world traffic benchmark datasets: PEMS03, PEMS04, PEMS07, and PEMS08 \cite{bib:others01:Chen}. Detailed descriptions of these datasets can be found in \tabref{tab:datasets}.

We partitioned each dataset using a 6:2:2 split for training, validation, and testing purposes, respectively. Traffic flows were aggregated at 5-minute intervals.
To provide a thorough evaluation of \sysname for both short-term and long-term traffic prediction, we examined four prediction scenarios where $T_h = T_p \in \{12, 48, 192, 288\}$.
Our performance metrics included the Root Mean Square Error (RMSE), Mean Absolute Error (MAE), and Mean Absolute Percentage Error (MAPE).

\subsection{Baselines and configurations}
\subsubsection{Baselines.}
We compare \sysname with following baselines: DLinear \cite{DBLP:conf/aaai/ZengCZ023}, Graph Wavenet \cite{bib:IJCAI19:Wu}, {AGCRN} \cite{DBLP:conf/nips/0001YL0020}, {STG-NCDE} \cite{bib:AAAI22:Choi}, {GMAN} \cite{DBLP:conf/aaai/ZhengFW020}, {SSTBAN} \cite{DBLP:conf/icde/0001LGWZSHW23}, {PDFormer} \cite{DBLP:conf/aaai/JiangHZW23}, and {STID} \cite{DBLP:conf/cikm/ShaoZ00X22}.
DLinear represents a linear-based approach for long time series forecasting.
Graph Wavenet is a notable CNN-based STGNN. Meanwhile, DCRNN, AGCRN, and STG-NCDE are RNN-based STGNNs.
GMAN, SSTBAN, and PDFormer are Transformer-based STGNNs. Notably, SSTBAN is the state-of-the-art STGNN for long-term traffic prediction, while PDFormer excels in short-term forecasting. STID is the state-of-the-art linear-based approach for traffic prediction.

\subsubsection{Configurations.}
For all datasets, models were trained over 300 epochs using the Adam optimizer \cite{bib:ICLR15:kingma}. The learning rate was set to 2e-4, with a batch size of 32. Hyperparameter $d$ was configured at 32, $e$ at 8, and $c$ at 32. The decoder has $L=3$ residual blocks. The moving average kernel size was optimized from among ${3, 5,15,25}$.
% \sysname was implemented in Pytorch and executed on NVIDIA A100 GPUs. For all baselines, we utilized their default configurations when running their codes. If experimental results were know for a scenario, we defaulted to their official results. 
% Otherwise, we
% It's noteworthy that certain results from PDFormer and GMAN were non-applicable due to computational constraints. While SSTBAN employed a unique data-preprocessing method, it failed to provide the PEMS03 and PEMS07 datasets processed using this method. Consequently, reproducing the results of SSTBAN on PEMS03 and PEMS07 using its source code proved impossible. We reported outcomes as the mean of five iterations.
All evaluations are implemented and executed on workstations equipped with NVIDIA A100 GPU. For all baseline models, we proceed as follows:
\begin{itemize}
    \item If experimental results for a specific scenario are not available, we execute their codes using default configurations.
    \item When official results are available, we directly cite these findings.
\end{itemize}
However, there are exceptions. The execution of PDFormer and GMAN's code on certain scenarios resulted in out-of-memory errors, rendering their results inapplicable for those scenarios. Furthermore, while SSTBAN employs a unique data-preprocessing technique, it does not furnish the processed PEMS03 and PEMS07 datasets. Consequently, we were unable to evaluate SSTBAN on these two datasets using its original source code. For our evaluations, we reported the average outcomes derived from five separate runs.

\subsection{Inference Accuracy}
In \tabref{tab:results}, we present an exhaustive evaluation of all baseline methods alongside \sysname, tested on four distinct datasets across forecast horizons of 12, 48, 192, and 288.
Our analysis leads to the following observations:
\begin{itemize}
    \item Across nearly all datasets and scenarios, \sysname consistently delivers top-tier performance. An exception to this trend is the PEMS08-12 scenario. Nonetheless, \sysname remains the leading model in the majority of scenarios for the PEMS08 dataset.
    %Specifically, compare to the second-best baseline PDFormer, \sysname has slight MAE\&MAPE increases on 12-horizons, but it reverses the situation on other scenarios, achieving better performance on all metrics. Furthermore, \sysname does not model the correlations among nodes at all, while PDFormer employs a spatial self-attention module to capture short- and long-range spatial dependencies simultaneously.
    
    % \item Transformer-based baseline approaches, including GMAN, {SSTBAN}, and PDformer, show superior performance amongst STGNNs, barring \sysname. 
    % This underscores the efficacy of the self-attention mechanism in capturing long-term dynamics.

    \item Compared to DLinear, STGNN-based methods display a pronounced edge. We hypothesize this advantage arises because STGNNs harness the power of GNNs to identify spatial interdependencies among traffic series. 
    Conversely, DLinear overlooks these spatial dependencies, making it less adept at handling the intricacies of traffic data. 
    On the contrary, however, \sysname outperforms STGNN-based baselines even without explicitly modeling these spatial dependencies. 
    This suggests the possibility that emphasizing spatial dependencies might not be quintessential. 
    Learning only spatial attributes could potentially serve as a more efficient alternative to GNNs.

    \item \sysname excels in both short-term and long-term traffic forecasting, outpacing competing methodologies. This highlights that even in the absence of sophisticated sequential models like Transformers, leveraging conventional time series decomposition techniques and embedding temporal patterns (time-of-the-day and day-of-the-week information) within simple linear layers can effectively discern the temporal correlations in traffic data.

\end{itemize} 

\begin{figure*}[t]   
  \centering            
  \subfloat[MACs/epoch with different prediction horizon length.]
  {
      \label{fig:subfig4}\includegraphics[width=0.41\textwidth]{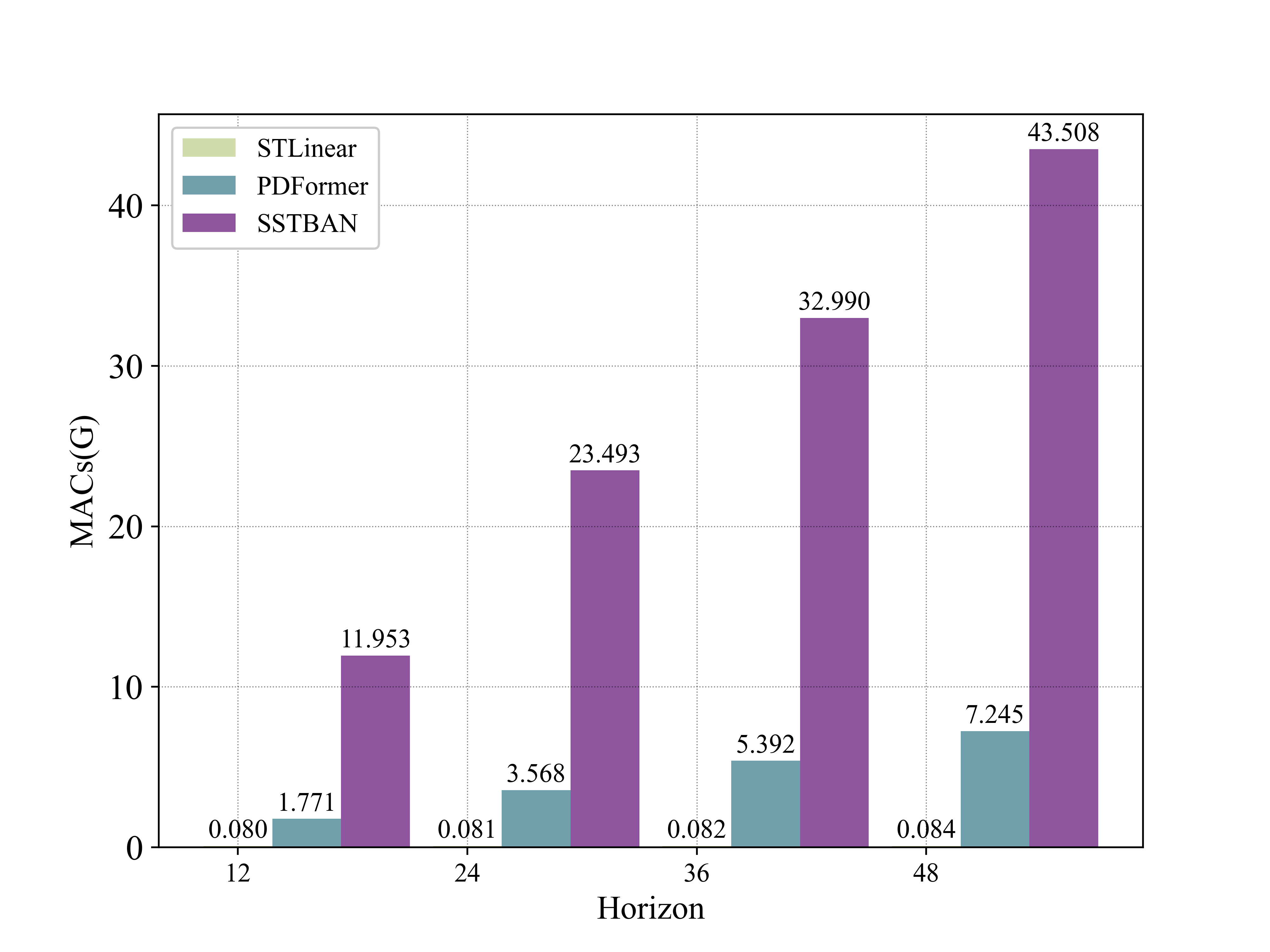}
  }
  \subfloat[Memory cost with different prediction horizon length.]
  {
      \label{fig:subfig5}\includegraphics[width=0.41\textwidth]{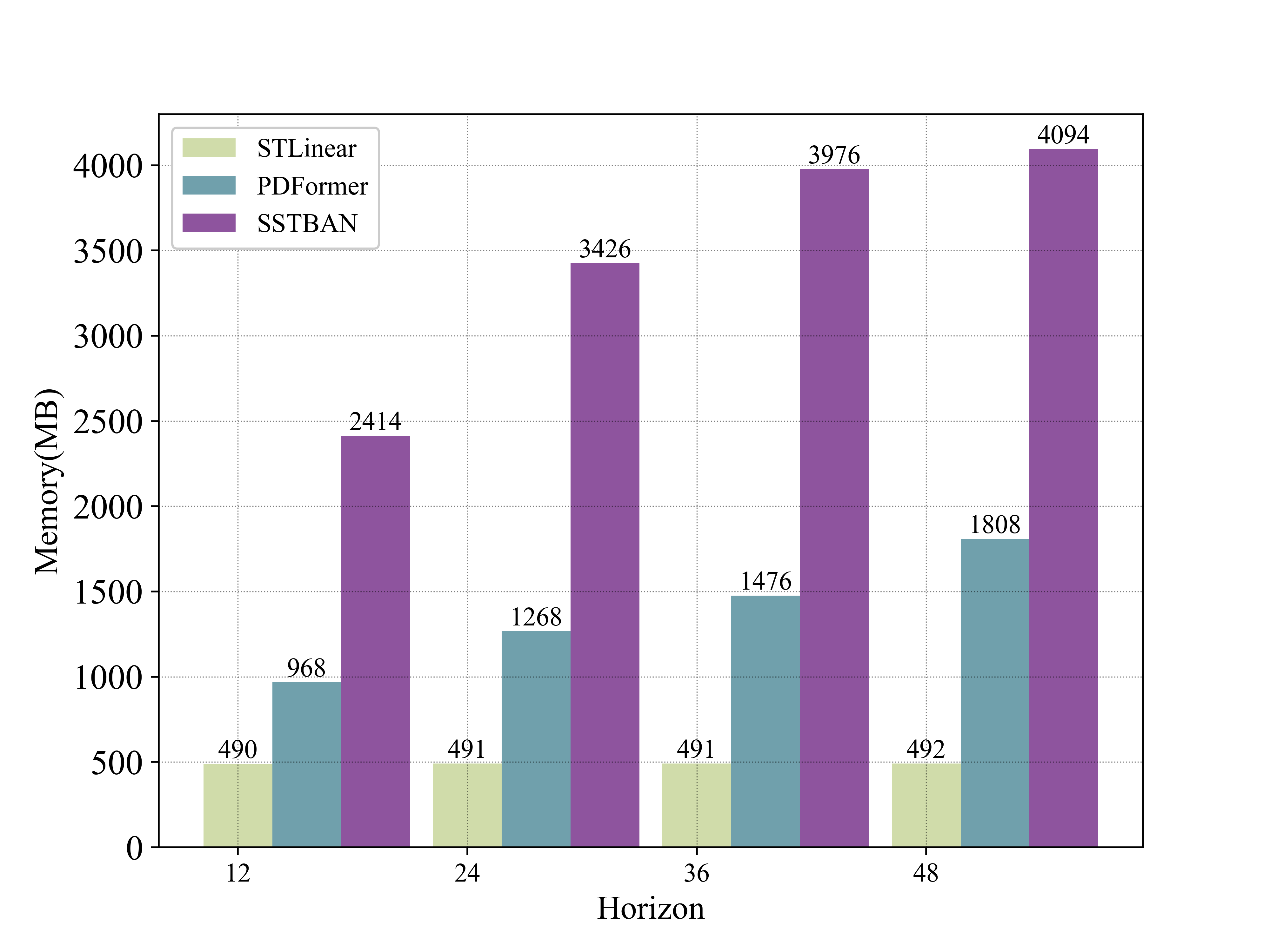}
  }

  \subfloat[MACs/epoch with different numbers of node.]
  {
      \label{fig:subfig6}\includegraphics[width=0.41\textwidth]{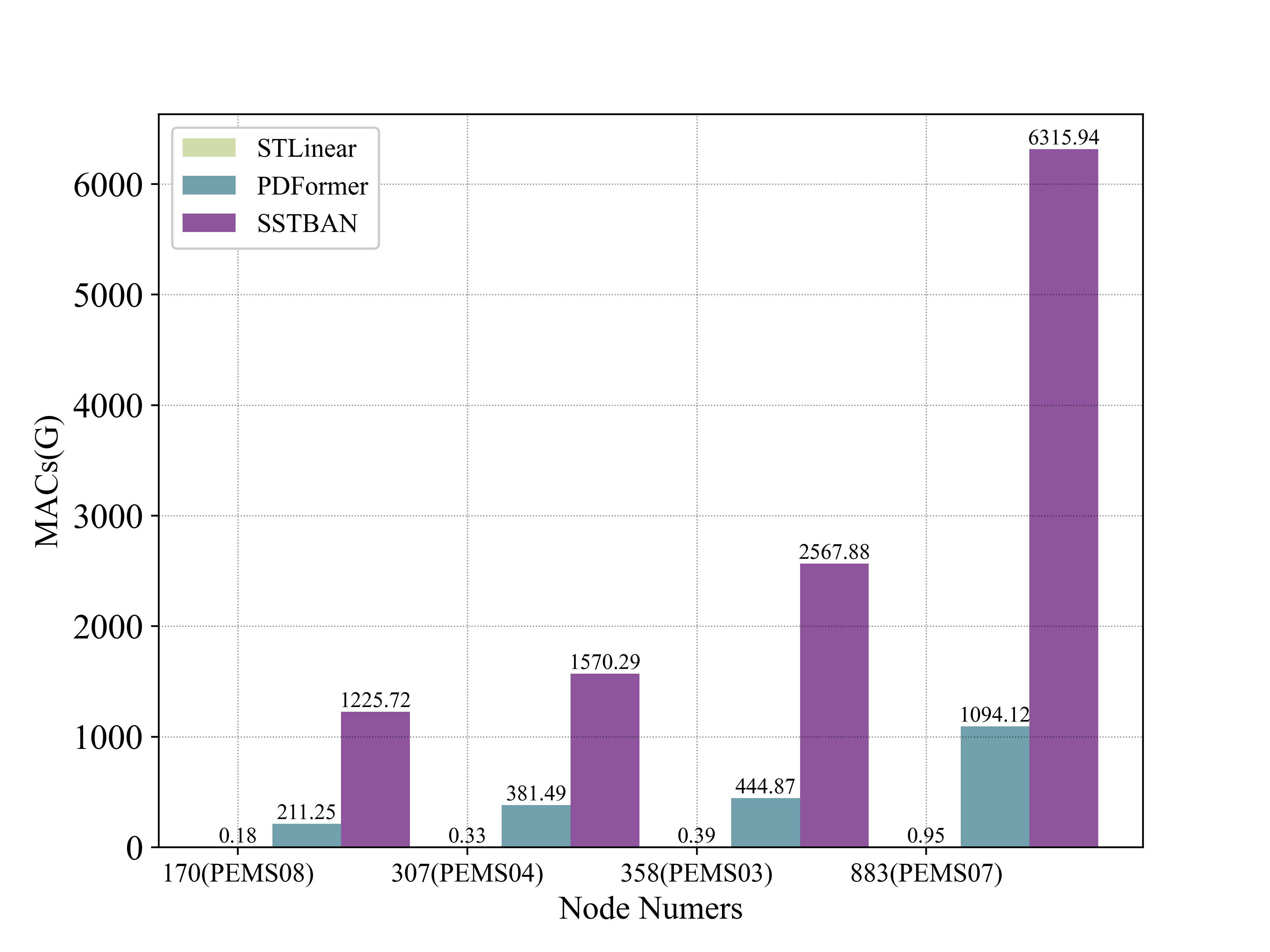}
  }
  \subfloat[Memory cost with different numbers of node.]
  {
      \label{fig:subfig7}\includegraphics[width=0.41\textwidth]{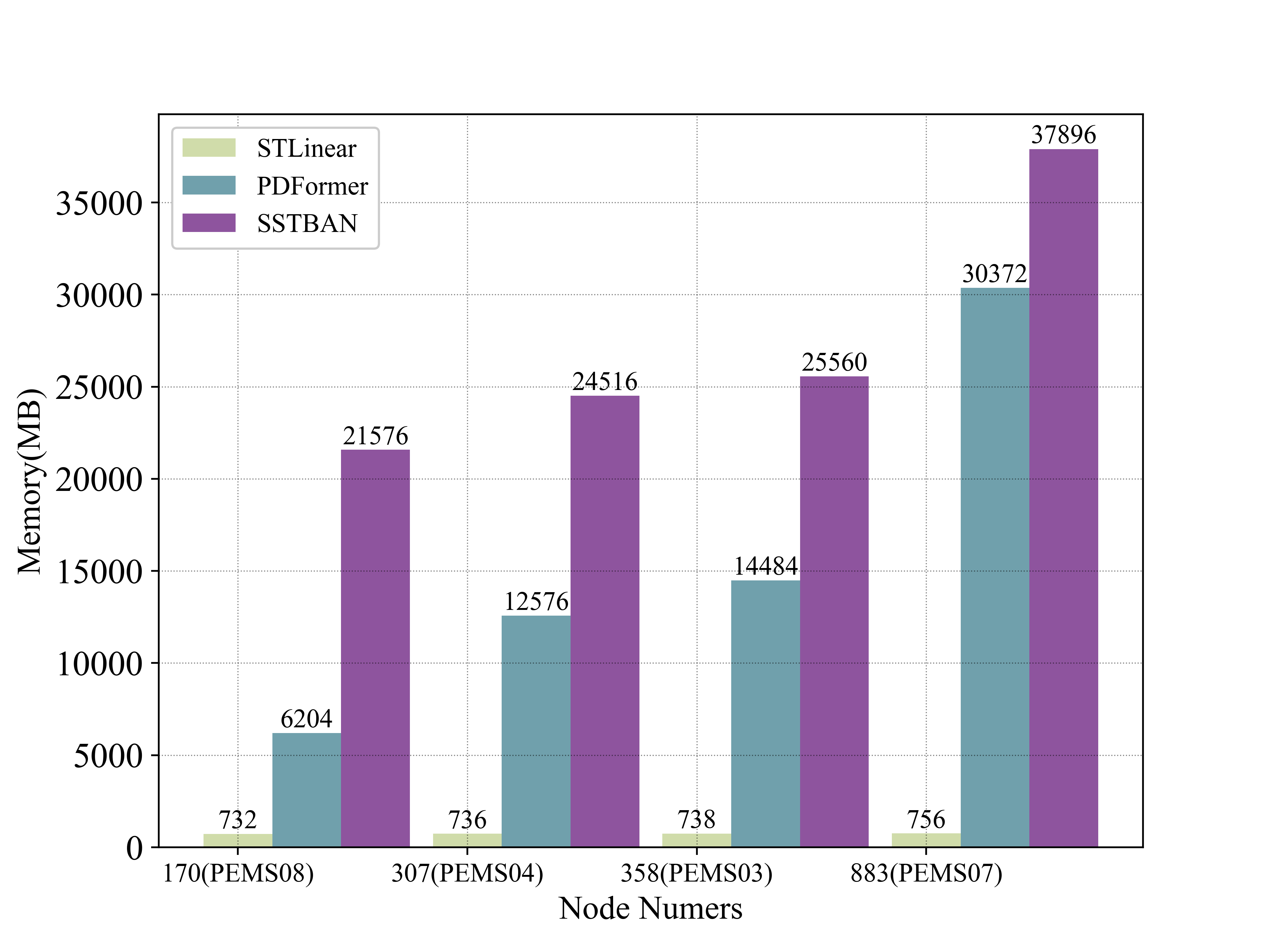}
  }

  \caption{(a) MACs per epoch for inference on dataset PEMS04 with prediction horizon length $\{12, 24, 36, 48\}$; (b) Memory cost for the same setup as (a); (c) MACs per epoch for inference on four different datasets containing 170, 307, 358 and 883 nodes, respectively. The prediction horizon length is fixed to 288; (d) Memory cost for the same setup as (c)}
  \label{fig:macs}            
\end{figure*}

\subsection{Efficiency}

\begin{table*}[htbp]

\centering
\caption{MACs per epoch and memory cost for training with prediction horizon length of $\{12, 48, 192, 288\}$.}
\scriptsize
\begin{tabular}{l|l|l|llll}
\Xhline{1pt}
Dataset                 & \multicolumn{2}{l|}{Method}               & 12             & 48             & 192             & 288              \\ \hline
\multirow{6}{*}{PEMS04} & \multirow{2}{*}{PDFormer} & MACs(G)/epoch & 4.64e+4($22.1\times$)  & 1.89e+5($86.7\times$)  & N/A             & N/A              \\
                        &                           & Memory(MB)    & 5.83e+3($2.6\times$)   & 2.32e+4($10.1\times$)  & OOM             & OOM              \\ \cline{2-7} 
                        & \multirow{2}{*}{SSTBAN}   & MACs(G)/epoch & 2.23e+5($105.9\times$) & 8.09e+5($371.0\times$) & N/A             & N/A              \\
                        &                           & Memory(MB)    & 4.31e+3($1.9\times$)   & 1.65e+4($7.2\times$)   & OOM             & OOM              \\ \cline{2-7} 
                        & \multirow{2}{*}{STLinear} & MACs(G)/epoch & 2.10e+3($1\times$)     & 2.18e+3($1\times$)     & 2.48e+3($1\times$)      & 2.68e+3($1\times$)       \\
                        &                           & Memory(MB)    & 2.27e+3($1\times$)     & 2.30e+3($1\times$)     & 2.43e+3($1\times$)      & 2.59e+3($1\times$)       \\ \hline
\multirow{6}{*}{PEMS08} & \multirow{2}{*}{PDFormer} & MACs(G)/epoch & 2.70e+4($22.1\times$)  & 1.10e+5($79.5\times$)  & 1.35e+5($93.6\times$)   & 6.57e+6(4209.7$\times$)  \\
                        &                           & Memory(MB)    & 2.61e+3($1.2\times$)   & 8.83e+3($4.0\times$)   & 2.71e+4($11.8\times$)   & 4.77e+4($20.5\times$)    \\ \cline{2-7} 
                        & \multirow{2}{*}{SSTBAN}   & MACs(G)/epoch & 1.99e+5($162.3\times$) & 6.64e+5($523.1\times$) & 3.37e+6($2232.6\times$) & 3.81e+7($24425.6\times$) \\
                        &                           & Memory(MB)    & 3.68e+3($1.7\times$)   & 8.93e+3($4.1\times$)   & 3.12e+4($13.6\times$)   & 4.84e+4($20.5\times$)    \\ \cline{2-7} 
                        & \multirow{2}{*}{STLinear} & MACs(G)/epoch & 1.22e+3($1\times$)     & 1.27e+3($1\times$)     & 1.45e+3($1\times$)      & 1.56e+3($1\times$)       \\
                        &                           & Memory(MB)    & 2.19e+3($1\times$)     & 2.20e+3($1\times$)     & 2.30e+3($1\times$)      & 2.32e+3($1\times$)       \\ \Xhline{1pt}
\end{tabular}
\label{tab:mac}
\end{table*}

Efficiency, one of the major motivation for the designing of \sysname, critically influences the practicality of deep-learning-enabled traffic prediction methods. \sysname has a time complexity of $\mathcal{O}(N)$. In other words, the computational expense of \sysname increases linearly with the number of nodes in the traffic network, far superior than the STGNN-based baselines with a quadratic time complexity $\mathcal{O}(N^2)$.

Apart from the theoretical time complexity, we empirically evaluated the training and inference efficiencies of \sysname against two state-of-the-art models: SSTBAN and PDFormer.
\tabref{tab:mac} provides a summary of the training efficiency on PEMS04 and PEMS08 datasets, while \figref{fig:macs} illustrates the influence of prediction horizon length $T_p$ and node count $N$ on inference efficiency.
Both during training and inference, \sysname stands out as the most efficient model. For example, in a direct comparison with SSTBAN (state-of-the-art method published in 2023), \sysname showcases a notable reduction in computational demands. Specifically, there's a reduction of 95.50\% to 99.81\% in MACs per epoch and 49.38\% to 98.01\% in memory usage. These efficiencies become even more evident during the training phase, with reductions ranging from 99.10\% to 99.96\% in MACs per epoch and 15.99\% to 95.20\% in memory usage.

It is pertinent to note that, during our evaluations on a platform equipped with 80 GB of graphic memory, both SSTBAN and PDFormer faced out-of-memory issues when tested on the PEMS04-192 and PEMS04-288 configurations. This observation underscores potential limitations in deploying Transformer-based models in extensive traffic prediction scenarios.
%As \figref{fig:macs} illustrates, the MACs of SSTBAN and PDFormer also grow linearly with the number of horizons and nodes, but STLinear achieves much lower MACs than both of them. Specifically, it uses less than 0.02\% MACs compared with previous SOTA method SSTBAN on the PEMS07-288, the largest benchmark dataset. 

\begin{figure*}[htbp]   
  \centering            
 
  \subfloat[MAE]
  {
      \label{fig:subfig1}\includegraphics[width=0.30\textwidth]{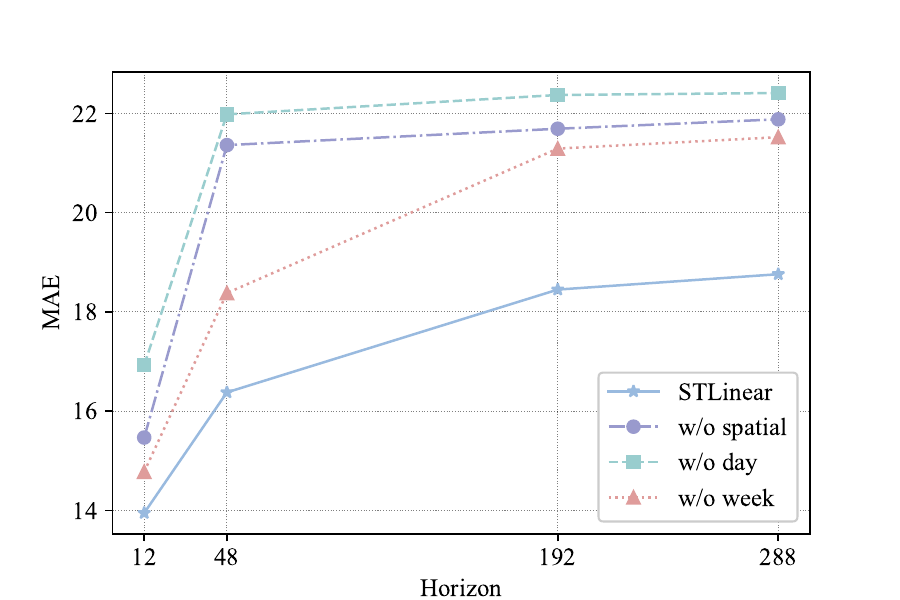}
  }
  \subfloat[RMSE]
  {
      \label{fig:subfig2}\includegraphics[width=0.30\textwidth]{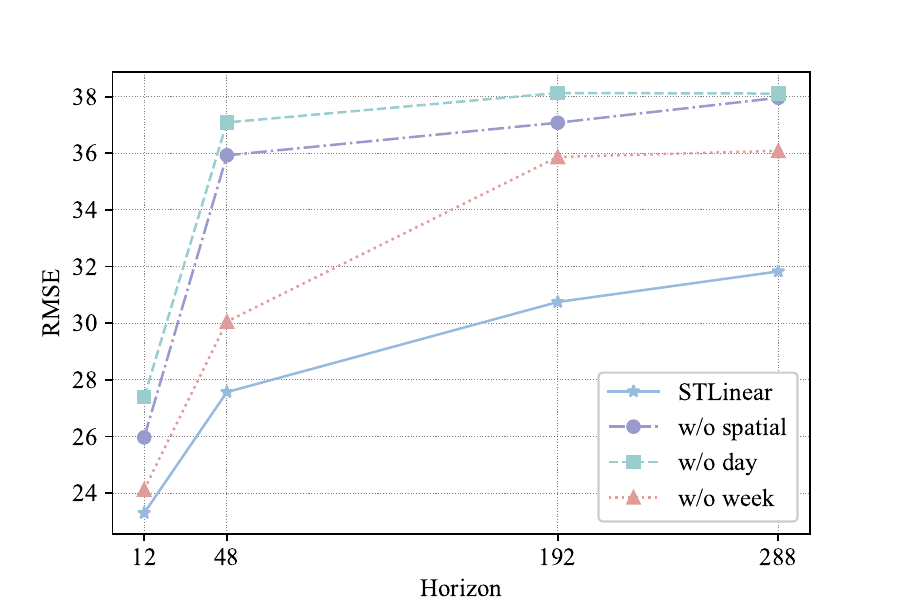}
  }
  \subfloat[MAPE]
  {
      \label{fig:subfig3}\includegraphics[width=0.30\textwidth]{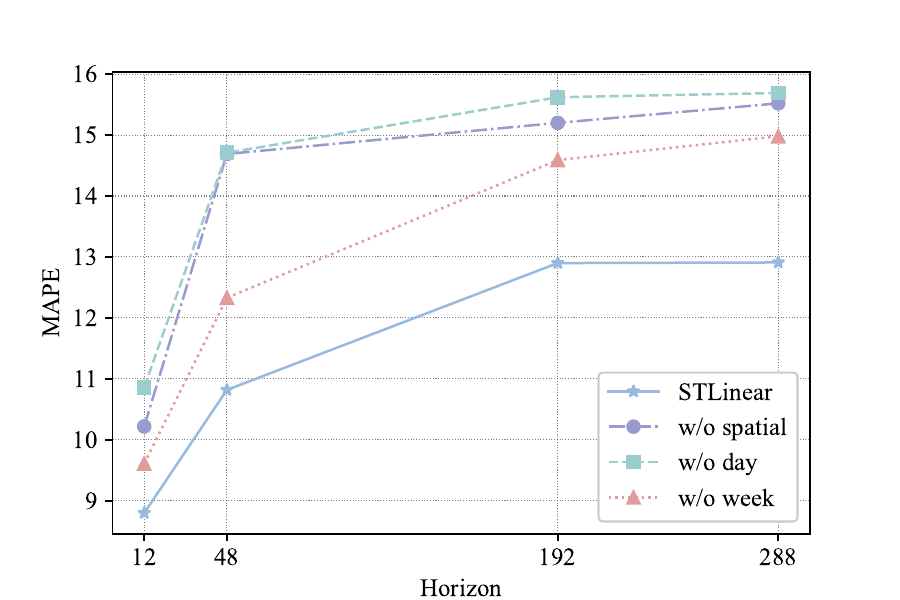}
  }
  \caption{Inference accuracy of \sysname against the three variants for the ablation study. Evaluated on PEMS08 dataset.}    
  \label{fig:ablation}            
\end{figure*}

\subsection{Contribution of Spatial Attributes and Periodicity}
\label{sec:ablation study}

In this section, we conduct ablation studies to validate the contributions of the three embedding vectors: \textit{(i)} the spatial embedding $\mathbf{s}_{i}$ modeling the spatial attributes; \textit{(ii)} the time-of-the-day embedding $\mathbf{d}(t)$; \textit{(iii)} the day-of-the-week embedding $\mathbf{w}(t)$. We assess \sysname against the following three model variants:

\begin{enumerate}
\item \textbf{w/o spatial embedding $\mathbf{s}_{i}$}: This variant omits the spatial embedding $\mathbf{s}_{i}$, implying that all nodes share the same parameters. Consequently, it cannot capture distinct spatial attributes associated with each node.
\item \textbf{w/o time-of-the-day embedding $\mathbf{d}(t)$}: This variant excludes the time-of-the-day embedding $\mathbf{d}(t)$ from both initial time vector $\mathbf{e}^{start}_{t}$ the terminal time vector $\mathbf{e}^{end}_{t}$.

\item \textbf{w/o day-of-the-week embedding $\mathbf{w}(t)$}: Similarly, this omits the day-of-the-week embedding $\mathbf{w}(t)$ from both initial time vector $\mathbf{e}^{start}_{t}$ the terminal time vector $\mathbf{e}^{end}_{t}$.

% \item \textbf{w/o initial time vector $\mathbf{e}^{start}_{t}$}: By removing the initial time vector, this variant prevents \sysname from accessing temporal span information from the input.

% \item \textbf{w/o terminal time vector $\mathbf{e}^{start}_{t}$}: Similarly, this variant excludes the terminal time vector $\mathbf{e}^{start}_{t}$.

\end{enumerate}

The comparative performance of these variants on the PEMS08 datasets is illustrated in \figref{fig:ablation}. From our findings, we infer the following:
\begin{itemize}
\item Overall, \sysname consistently achieves better results than all other variants, showing that the inclusion of the spatial attributes learning and periodicity information indeed enhances the model's capabilities.

\item In every scenario, the absence of the time-of-the-day embedding adversely affects the performance of \sysname. The \textbf{w/o time-of-the-day embedding} variant is consistently outperformed by other models, emphasizing that daily periodicity is a critical factor in traffic prediction.

\item The \textbf{w/o spatial embedding} variant performs better than the \textbf{w/o time-of-the-day embedding} but falls short of the \textbf{w/o day-of-the-week embedding}. This suggests that the inability to learn spatial attributes has a more pronounced impact than the absence of weekly periodicity, but it's less detrimental than missing daily periodicity.

% \item \sysname consistently achieves better results than all other variants. This underscores the importance of recognizing the entire temporal span over identifying just the start or end times.

\end{itemize}

%% file: body/conclusion.tex
\section{Conclusion}

Our study sought to address the limitations of the application of state-of-the-art STGNNs on traffic prediction tasks. Driven by recent findings, we pinpointed and advocated for three primary solutions to enhance efficiency: the emphasis on a node-embedding approach, the endorsement of time series decomposition, and the introduction of periodicity learning. To actualize these solutions, we introduced \sysname, a minimalist and innovative model architecture. Distinguished by its fully localized nature, \sysname eliminates the need for inter-node data exchanges during its operations, thus reducing computational overhead and providing a pathway for deployment in communication-restricted environments. 
The model's reliance on pure linear layers further amplifies its computational efficiency.

Empirical analyses conducted on real-world datasets underscore the efficacy and efficiency of \sysname. Notably, it matches or surpasses established STGNNs in traffic prediction accuracy while drastically curtailing computational demands. Compared to the benchmark SSTBAN method, \sysname exhibited significant reductions in MACs per epoch during both training and inference stages, coupled with reduced memory consumption. In essence, our work underscores the potential of \sysname to serve as a streamlined, yet potent alternative to traditional STGNNs, offering a more efficient path forward for ITS and smart city development.
In the larger canvas of traffic prediction research, our findings may potentially allow a paradigm shift. With models like \sysname, it is feasible to have the best of both worlds: robust predictive capabilities with streamlined computational requirements. As urban landscapes continue to evolve and as the demands on ITS grow, solutions like \sysname will prove invaluable in bridging the gap between theoretical advancements and practical implementations in real-world traffic management systems.